\documentclass[conference]{IEEEtran}
\IEEEoverridecommandlockouts
\usepackage{cite}
\usepackage{amsmath,amssymb,amsfonts}
\usepackage{algorithm}
\usepackage{algorithmicx}
\usepackage{algpseudocode}
\usepackage{amsmath}
\usepackage{graphicx}
\usepackage{textcomp}
\usepackage{xcolor}
\usepackage{booktabs}
\usepackage{tabularx}
\usepackage{makecell}
\usepackage{pifont}
\usepackage{wasysym} 
\usepackage[table]{xcolor}
\usepackage{colortbl}
\usepackage[most]{tcolorbox}

\usepackage{microtype}

\usepackage{xcolor}
\usepackage{subcaption} 

\usepackage{hyperref}
\usepackage{ragged2e}    
\usepackage{seqsplit} 
\usepackage{booktabs,multirow,makecell}
\renewcommand{\arraystretch}{1.1}
\usepackage{kotex}
\newcommand{\concat}{\mathbin{\|}} 
\newcolumntype{Y}{>{\raggedright\arraybackslash}X}
\usepackage{listings}

\usepackage{siunitx}
\def\BibTeX{{\rm B\kern-.05em{\sc i\kern-.025em b}\kern-.08em
    T\kern-.1667em\lower.7ex\hbox{E}\kern-.125emX}}
\begin{document}

\title{Relation-Aware Bayesian Optimization of DBMS Configurations Guided by Affinity Scores*\\
}

\author{\IEEEauthorblockN{Sein Kwon}
\IEEEauthorblockA{\textit{Computer Science} \\
\textit{Yonsei Univercity}\\
Seoul, Korea \\
seinkwon97@yonsei.ac.kr}
\and
\IEEEauthorblockN{Seulgi Baek}
\IEEEauthorblockA{\textit{Computer Science} \\
\textit{Yonsei Univercity}\\
Seoul, Korea \\
seulgibaek@yonsei.ac.kr}
\and
\IEEEauthorblockN{Hyunseo Yang}
\IEEEauthorblockA{\textit{Computer Science} \\
\textit{Yonsei Univercity}\\
Seoul, Korea \\
hseo0608@yonsei.ac.kr}
\and
\IEEEauthorblockN{Youngwan Jo}
\IEEEauthorblockA{\textit{Computer Science} \\
\textit{Yonsei Univercity}\\
Seoul, Korea \\
jyy1551@yonsei.ac.kr}
\and
\IEEEauthorblockN{Sanghyun Park}
\IEEEauthorblockA{\textit{Computer Science} \\
\textit{Yonsei Univercity}\\
Seoul, Korea \\
sanghyun@yonsei.ac.kr}
}

\maketitle

\begin{abstract}
Database Management Systems (DBMSs) are fundamental for managing large-scale and heterogeneous data, and their performance is critically influenced by configuration parameters. Effective tuning of these parameters is essential for adapting to diverse workloads and maximizing throughput while minimizing latency. Recent research has focused on automated configuration optimization using machine learning; however, existing approaches still exhibit several key limitations.
Most tuning frameworks disregard the dependencies among parameters, assuming that each operates independently. This simplification prevents optimizers from leveraging relational effects across parameters, limiting their capacity to capture performance-sensitive interactions. Moreover, to reduce the complexity of the high-dimensional search space, prior work often selects only the top few parameters for optimization, overlooking others that contribute meaningfully to performance. Bayesian Optimization (BO), the most common method for automatic tuning, is also constrained by its reliance on surrogate models, which can lead to unstable predictions and inefficient exploration.
To overcome these limitations, we propose RelTune, a novel framework that represents parameter dependencies as a Relational Graph and learns GNN-based latent embeddings that encode performance-relevant semantics. RelTune further introduces Hybrid-Score-Guided Bayesian Optimization (HBO), which combines surrogate predictions with an Affinity Score measuring proximity to previously high-performing configurations. Experimental results on multiple DBMSs and workloads demonstrate that RelTune achieves faster convergence and higher optimization efficiency than conventional BO-based methods, achieving state-of-the-art performance across all evaluated scenarios.
  
\end{abstract}


\section{Introduction}
In recent years, the exponential growth of data has made the performance of database management systems (DBMSs) increasingly critical for efficiently storing and processing large-scale data \cite{DB_survey}. To enhance DBMS performance, various optimization approaches have been actively studied  \cite{Tuning_survey}. Among them, database parameter tuning aims to optimize performance by appropriately adjusting the numerous configuration parameters within the DBMS \cite{DB_review}. Traditionally, a Database Administrator (DBA) manually tuned these parameters in a heuristic manner to achieve better performance. However, this manual process has several limitations. First, since modern DBMSs contain a large and diverse set of parameters, it is difficult for a DBA to fully understand all parameters and their complex interactions with system performance. Moreover, because different DBMSs have different types and semantics of parameters, it is challenging for a DBA to consider all possible combinations across systems. To address these challenges, recent research has explored automatic database parameter tuning using machine learning (ML) techniques, which automatically optimize database parameters based on observed configuration and performance data \cite{iTuned,Tuneful,OtterTune,ResTune,CDBTune,Qtune,Facilitating, csat, gptuner}. Although existing ML-based approaches, such as those leveraging Bayesian Optimization (BO) \cite{BO} or Reinforcement Learning (RL) \cite{RL}, have shown high efficiency compared to heuristic search, they still suffer from several fundamental limitations. 

One persistent limitation of existing studies is the neglect of dependencies among configuration parameters. Most existing approaches assume parameter independence and ignore inter-parameter relationships during optimization.

\begin{figure}[t]
  \centering
  \begin{minipage}{0.485\columnwidth}
    \centering
    \includegraphics[width=\linewidth]{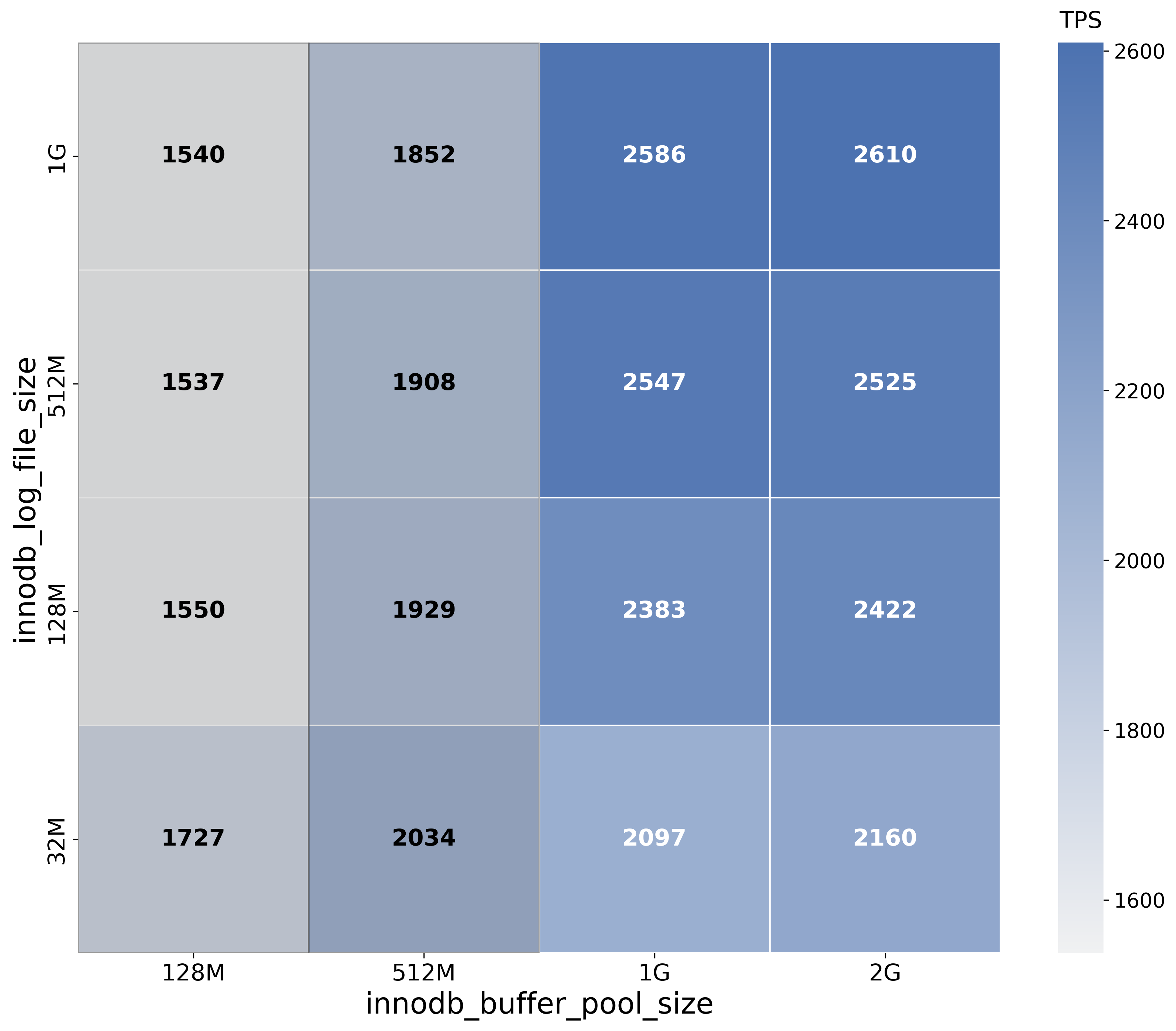}
  \end{minipage}\hfill
  \begin{minipage}{0.485\columnwidth}
    \centering
    \includegraphics[width=\linewidth]{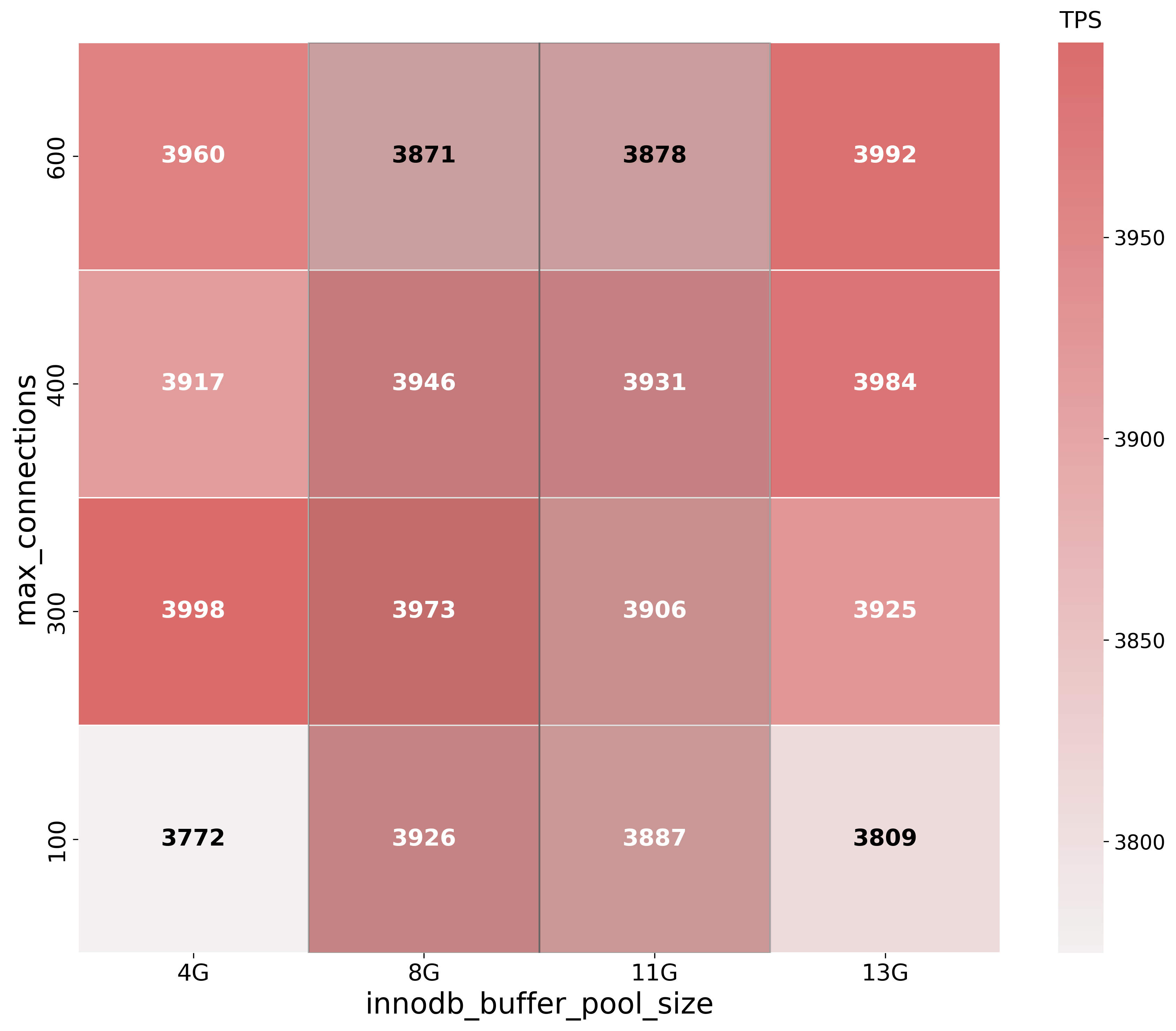}
  \end{minipage}
  \caption{Throughput comparison of correlated (left) and independent (right) parameter pairs.}
  \label{fig:intro_ex}
\end{figure}

However, in practice, many parameters are interdependent, exhibiting conditional constraints or value-related dependencies as documented in official DBMS manuals \cite{mysql_manual}. For example, the MySQL Manual states that \textit{“The performance effect of innodb\_buffer\_pool\_instances is only significant when innodb\_buffer\_pool\_size is large enough.”} indicating that the influence of one parameter is conditionally dependent on the level of another. Table \ref{tab:mysql-dependencies} summarized multiple examples of such parameter dependencies described in the MySQL 5.7 Manual, demonstrating that these relationships are difficult to capture under optimization frameworks that assume parameter independence. Figure \ref{fig:intro_ex} compares the throughput performance of MySQL under two different parameter combinations. 
The first heatmap (left) illustrates the interaction between \texttt{innodb\_buffer\_pool\_size} and \texttt{innodb\_log\_file\_size},  revealing a clear conditional dependency between the two parameters. For instance, when \texttt{innodb\_log\_file\_size} is fixed at 128 MB, increasing \texttt{innodb\_buffer\_pool\_size} from 1 GB to 2 GB improves throughput, whereas under a different condition where \texttt{innodb\_log\_file\_size} is set to 512 MB, the same increase results in performance degradation. 

In contrast, the right heatmap shows the combination of \texttt{innodb\_buffer\_pool\_size} and \texttt{max\_connections}. In this case, throughput remains nearly constant across different configurations, 
and no monotonic trend is observed along either parameter axis. The variation range is small, indicating that changes in one parameter do not noticeably influence the effect of the other. This suggests that these two parameters act largely independently without significant interaction. 

This comparison demonstrates that database parameters can operate both independently and conditionally dependently, depending on their functional relationships. Consequently, optimization methods assuming parameter independence may overlook critical global optima, highlighting the importance of explicitly modeling parameter relationships for reliable and efficient configuration tuning.

\newcolumntype{S}[1]{>{\raggedright\arraybackslash\scriptsize}p{#1}}
\newcolumntype{Y}{>{\raggedright\arraybackslash}X}
\begin{table}[t]
\centering
\setlength{\tabcolsep}{4pt}
\renewcommand{\arraystretch}{1.12}
\footnotesize
\caption{Representative MySQL parameter dependencies. All parameters refer to InnoDB system variables.}
\label{tab:mysql-dependencies}

\rowcolors{2}{white}{gray!6}
\begin{tabularx}{\linewidth}{S{23mm} S{26mm} Y} 
\toprule
\textbf{Param A} & \textbf{Param B} & \textbf{Manual Evidence} \\
\midrule
buffer\_pool\_size & buffer\_pool\_instances
& ``... The number of buffer pool instances has no effect \textcolor{blue!70!black}{unless the pool is large}.'' \\

io\_capacity & max\_dirty\_pages\_pct
& ``... The rate at which InnoDB flushing rate \textcolor{blue!70!black}{depends on I/O capacity}.'' \\

doublewrite & flush\_method
& ``... Doublewrite can bottleneck \textcolor{blue!70!black}{when using O\_DIRECT}.'' \\
\bottomrule
\end{tabularx}

\vspace{2pt}
\scriptsize Sources: MySQL 5.7 manual.
\end{table}

Another limitation of existing studies lies in their handling of the curse of dimensionality in high-dimensional optimization \cite{bo_highdimen, bo_highdimen_2}. To alleviate this issue, prior studies typically avoid optimizing the entire configuration space. Instead, they identify and tune only a subset of parameters that are believed to have the greatest impact on performance, often selected using algorithms such as LASSO \cite{LASSO} or SHapley Additive exPlanations (SHAP) \cite{SHAP}. 
However, such feature selection based approaches may ignore valuable parameters that are meaningful but less dominant. Moreover, the selected parameters often differ across algorithms, resulting in inconsistency. Table \ref{tab:top-10-mysql} compares the top-10 parameters identified by LASSO and SHAP for two MySQL workloads.
The results show that only three or five parameters overlap between the two algorithms for each workload.
This inconsistency suggests that the definition of top parameters is highly sensitive to the workload characteristics and the feature selection method. Consequently, the final optimized configurations derived from these different parameter subsets can lead to noticeably different performance outcomes, undermining the reliability of partial-parameter optimization.

A further limitation arises from the inherent characteristics of BO, which is fundamentally a black-box optimization method that heavily depends on the performance of its surrogate model \cite{bo_surrogate, bo_surrogate_2}. This dependency becomes particularly problematic in noisy or high-dimensional search spaces, where the surrogate tends to produce unreliable predictions, ultimately leading to inefficient optimization outcomes. Moreover, this issue often leads to instability during the early exploration phase, causing the surrogate to repeatedly explore irrelevant regions of the search space, thereby reducing overall optimization efficiency.


To address the aforementioned limitations, we propose \textbf{RelTune}, a relation-aware Bayesian Optimization framework that integrates parameter dependencies and affinity-based guidance for efficient DBMS configuration tuning. To overcome the limitation of prior studies that overlooked such parameters relationships, RelTune leverages a Large Language Model (LLM) to extract descriptions of parameters, covering functionalities and interactions, from official DBMS documentation. Based on these descriptions, RelTune constructs a parameter dependency graph that captures complex relationships among parameters, thereby enabling the use of domain knowledge during the tuning process. To further mitigate the curse of dimensionality and the information loss caused by optimizing only a small subset of parameters, RelTune employs a Graph Neural Network (GNN) \cite{GNN} to learn representations from the constructed graph. The GNN encodes the relationships among parameters and compresses them into a low-dimensional latent space. This latent representation not only reduces dimensionality but also preserves the structural dependencies among parameters, enabling efficient optimization across the entire configuration space. Additionally, to overcome the exploration inefficiency of conventional BO, we introduce an Affinity Score that enhances the acquisition process by incorporating the structural information of the latent space. This score guides the search toward regions that are close to previously high-performing latent vectors, thereby concentrating the optimization process on promising areas. We evaluate RelTune on two DBMSs, MySQL \cite{mysql} and PostgreSQL \cite{Postgres}, across four and two workloads, respectively. Experimental results demonstrate that RelTune achieves substantial performance improvements over state-of-the-art methods.

\begin{table*}[!htp]
\centering

\caption{Top-10 parameters using SHAP and LASSO for MySQL Read50Update50, Read95Update5 workloads. Overlapped parameters are highlighted in bold.}
\label{tab:top-10-mysql}
\fontsize{10pt}{12pt}\selectfont
\renewcommand{\arraystretch}{1.3}
\hspace*{-0.2\textwidth}
\resizebox{0.8\textwidth}{!}{ 
\begin{tabular}{c c}
    \begin{minipage}{0.4\linewidth}
    \centering
    \begin{tabular}{clcl}
    \hline
    \multicolumn{4}{c}{\textbf{ Read50Update50}} \\ \hline
    \multicolumn{2}{c}{\textbf{SHAP (top-10)}}                      & \multicolumn{2}{c}{\textbf{Lasso (top-10)}}             \\ \hline

\multicolumn{2}{l}{\textbf{range\_alloc\_block\_size}}          & \multicolumn{2}{l}{\textbf{range\_alloc\_block\_size}}  \\
\multicolumn{2}{l}{\textbf{innodb\_buffer\_pool\_size}}         & \multicolumn{2}{l}{\textbf{innodb\_buffer\_pool\_size}} \\
\multicolumn{2}{l}{\textbf{innodb\_spin\_wait\_delay}}          & \multicolumn{2}{l}{\textbf{innodb\_spin\_wait\_delay}}  \\
\multicolumn{2}{l}{innodb\_purge\_threads}                      & \multicolumn{2}{l}{\textbf{innodb\_stats\_persistent}} \\
\multicolumn{2}{l}{\textbf{innodb\_log\_file\_size}}             & \multicolumn{2}{l}{optimizer\_prune\_level}             \\
\multicolumn{2}{l}{\textbf{innodb\_stats\_persistent}}          & \multicolumn{2}{l}{innodb\_optimize\_filltext\_only}     \\
\multicolumn{2}{l}{innodb\_old\_blocks\_time}                    & \multicolumn{2}{l}{innodb\_read\_io\_threads}            \\
\multicolumn{2}{l}{innodb\_ft\_min\_token\_size}                  & \multicolumn{2}{l}{\textbf{innodb\_log\_file\_size}}     \\
\multicolumn{2}{l}{binlog\_cache\_size}                         & \multicolumn{2}{l}{max\_error\_count}                   \\
\multicolumn{2}{l}{inodb\_compression\_failure\_threshold\_pct}   & \multicolumn{2}{l}{innodb\_ft\_max\_token\_size}          \\ \hline

    \end{tabular}
    \end{minipage}
    &
    \hspace{4cm}
    \begin{minipage}{0.4\linewidth}
    \centering
    \begin{tabular}{clcl}
    \hline
    \multicolumn{4}{c}{\textbf{ Read95Update5}} \\ \hline
    \multicolumn{2}{c}{\textbf{SHAP (top-10)}}               & \multicolumn{2}{c}{\textbf{Lasso (top-10)}}               \\ \hline

\multicolumn{2}{l}{\textbf{range\_alloc\_block\_size}}   & \multicolumn{2}{l}{\textbf{range\_alloc\_block\_size}}    \\
\multicolumn{2}{l}{\textbf{innodb\_buffer\_pool\_size}}  & \multicolumn{2}{l}{\textbf{innodb\_spin\_wait\_delay}}    \\
\multicolumn{2}{l}{innodb\_spin\_wait\_delay}            & \multicolumn{2}{l}{innodb\_optimize\_fulltext\_only}      \\
\multicolumn{2}{l}{\textbf{innodb\_random\_read\_ahead}} & \multicolumn{2}{l}{innodb\_file\_per\_table}              \\
\multicolumn{2}{l}{innodb\_purge\_threads}               & \multicolumn{2}{l}{binlog\_cache\_size}                   \\
\multicolumn{2}{l}{sort\_buffer\_size}                   & \multicolumn{2}{l}{innodb\_print\_all\_deadlocks}         \\
\multicolumn{2}{l}{innodb\_read\_ahead\_threshold}       & \multicolumn{2}{l}{\textbf{innodb\_random\_read\_ahead}}  \\
\multicolumn{2}{l}{innodb\_lru\_scan\_depth}             & \multicolumn{2}{l}{innodb\_read\_io\_threads}             \\
\multicolumn{2}{l}{table\_open\_cache}                   & \multicolumn{2}{l}{query\_cache\_wlock\_invalidate}       \\
\multicolumn{2}{l}{innodb\_adaptive\_hash\_index\_parts}  & \multicolumn{2}{l}{innodb\_online\_alter\_log\_max\_size}   \\
 \hline
    \end{tabular}
    \end{minipage}
\end{tabular}
}
\end{table*}

Our main contributions are summarized as follows:
\begin{itemize}
  \item We construct a relational graph that captures structural dependencies among DBMS parameters based on descriptions generated by LLMs.  
  
  \item We train a GNN to jointly encode parameter relations and performance metrics into a compact latent space, enabling holistic optimization over all parameters. 
  
  \item We propose a Hybrid Score-Guided Bayesian Optimization that incorporates an Affinity Score to enhance exploration efficiency and tuning stability.
  
  \item Experimental results demonstrate that RelTune achieves superior performance through comparison with various baseline models.
\end{itemize}

\section{preliminaries}
\subsection{Automatic Database Parmeter Tuning}
Database parameter tuning aims to maximize the performance of a DBMS by optimizing its configuration parameters. Let the set of parameters be denoted as $P$, where $P_1, \dots, P_k$ represent $K$ parameters.
A collection of these parameters, $conf = {P_1, \dots, P_k}$, is defined as a configuration. The objective function is defined based on two key DBMS performance metrics: throughput and latency. Throughput measures the amount of work completed per unit time and should be maximized, while latency represents the delay per operation and should be minimized. Automatic parameter tuning approaches can be broadly categorized into search-based methods \cite{bestconfig} and ML-based methods \cite{iTuned, Tuneful, OtterTune, ResTune, CDBTune, Qtune, Facilitating, csat, gptuner}.

Search-based methods aim to identify optimal database configurations by following predefined rules or heuristics. A representative approach is BestConfig \cite{bestconfig}, which collects configuration and metric pairs and employs a sampling strategy called DDS (Divide and Diverge Sampling) to efficiently explore the high-dimensional parameter space while maintaining broad applicability. After identifying the configuration that yields the best performance, BestConfig further refines the search by proposing new parameter settings within a bounded region around the current optimum using the RBS (Recursive Bound and Search) algorithm. However, such rule-based tuning approaches are inherently data dependent. Their performance can become unstable when the collected data are noisy or limited in quantity. In addition, since they rely on heuristic exploration rather than learned models, these methods often require substantial time to converge to an optimal configuration.

ML-based methods leverage learned models to predict performance and identify optimal parameter configurations. A representative approach is OtterTune \cite{OtterTune}, which stores configuration and metric pairs generated during the optimization process in a data repository. These data are then used during the workload mapping stage to enable optimization across different workloads. To alleviate the inefficiency of exploring high-dimensional spaces, OtterTune employs LASSO to select the most influential parameters and trains a Gaussian Process (GP)–based \cite{GP} surrogate model only on this selected subset. However, since only a subset of parameters is optimized, valuable parameters may be omitted, and parameter relationships are not considered. Moreover, as OtterTune relies on BO, its performance remains highly dependent on the predictive accuracy of the surrogate model.

CDBTune \cite{CDBTune} adopts a RL approach that learns database configurations through a try-and-error process. To enable optimization in high-dimensional parameter spaces, CDBTune employs the Deep Deterministic Policy Gradient (DDPG) algorithm, which combines the advantages of the Deep Q-Network (DQN) \cite{DQN} and Actor–Critic \cite{actor-critic} methods. However, RL-based tuning methods such as CDBTune require a large amount of high-quality data and numerous iterations to achieve convergence. Moreover, since each configuration must be evaluated through actual benchmarking, the overall tuning process becomes time-consuming and computationally expensive.

RGPE (Ranking-weighted Gaussian Process Ensemble) \cite{Facilitating} represents one of the most effective algorithms among existing ML-based database tuning approaches, as demonstrated in comparative studies evaluating various methodologies such as knowledge transfer and workload mapping. RGPE is an ensemble model designed for BO-based optimization, which combines both workload-specific information and historical knowledge to guide the search process. However, similar to OtterTune, RGPE is unable to efficiently handle high-dimensional parameter spaces and therefore performs optimization only on a selected subset of parameters after a parameter selection step. In addition, the model requires careful tuning of hyper-parameters, which further increases the complexity of the optimization process.

CSAT (Confidence-Aware Surrogate-Assisted Tuning) \cite{csat} is a BO-based framework designed to improve the efficiency and stability of optimization by incorporating the confidence of the surrogate model.
To address the issue in conventional BO methods that often ignore surrogate uncertainty and consequently explore unreliable regions, CSAT introduces an acquisition function that integrates the surrogate’s confidence, thereby guiding the search toward more reliable regions of the parameter space.
This approach enhances the stability of exploration, particularly under noisy environments.
However, CSAT still fails to explicitly capture the inter-dependencies among parameters.
While it contributes to improving the reliability of the surrogate model, it does not account for the structural relationships or dependencies among configurations, making it incapable of leveraging relational knowledge for more efficient exploration.

GPTuner \cite{gptuner} is an LLM-based automatic database tuning framework that aims to overcome the limitations of conventional ML-based methods, which fail to fully leverage domain knowledge.
To achieve this, GPTuner collects and refines domain knowledge using a LLM (GPT-4) \cite{GPT} and transforms it into a structured representation that can be utilized during the optimization process.
It further introduces a Coarse-to-Fine BO strategy, which progressively explores a knowledge-guided, reduced search space to achieve efficient optimization.
While GPTuner effectively reduces tuning cost by structurally incorporating domain knowledge through LLMs, it still does not explicitly model the dependencies among parameters.
Specifically, GPTuner relies on LLM-generated explanatory, parameter-wise guidelines, but it does not represent parameter interactions in a graph-based form or learn the structural correlations within the search space.

\subsection{Bayesian Optimization}


BO \cite{BO}is widely used for black-box function optimization and is particularly effective for problems with expensive evaluation costs, such as database parameter tuning.
BO learns a surrogate model based on the observed configuration–metric pairs $(X, Y)$ to approximate the underlying objective function.
An acquisition function, denoted as $a(x)$, is then used to select the next candidate configuration for evaluation.
In most cases, a GP \cite{GP} is adopted as the surrogate model, which models the objective function as a probabilistic distribution defined by a mean function $\mu(x)$ and a covariance (kernel) function $k(x, x')$.
This formulation allows the GP to estimate both the expected value and the uncertainty of each configuration.
The acquisition function utilizes this predictive distribution to balance exploration and exploitation.
Typical acquisition functions, such as Expected Improvement (EI) \cite{EI} and Upper Confidence Bound (UCB) \cite{UCB}, use this predictive distribution to guide the selection of the next configuration $x_t$, which is determined as follows:

\begin{equation}
x_t = \arg\max_{x \in X} a(x)
\end{equation}
At each iteration, BO updates the surrogate model and selects a new candidate configuration to gradually move toward the global optimum. The dataset is then updated as
\begin{equation}
D_t = D_{t-1} \cup \{ (x_t, y_t) \},
\end{equation}
where $x_t$ and $y_t$ denote the newly explored configuration and its corresponding performance metric, respectively.
However, BO is highly sensitive to the predictive accuracy of its surrogate model. When the surrogate model fails to accurately approximate the underlying objective function, owing to limited training samples, measurement noise, or irregular performance landscapes, the acquisition function can lead the optimization toward suboptimal or unstable regions. These limitations directly motivate the design of RelTune.

\begin{figure*}[t]
    \centering
    \includegraphics[width=0.70\linewidth]{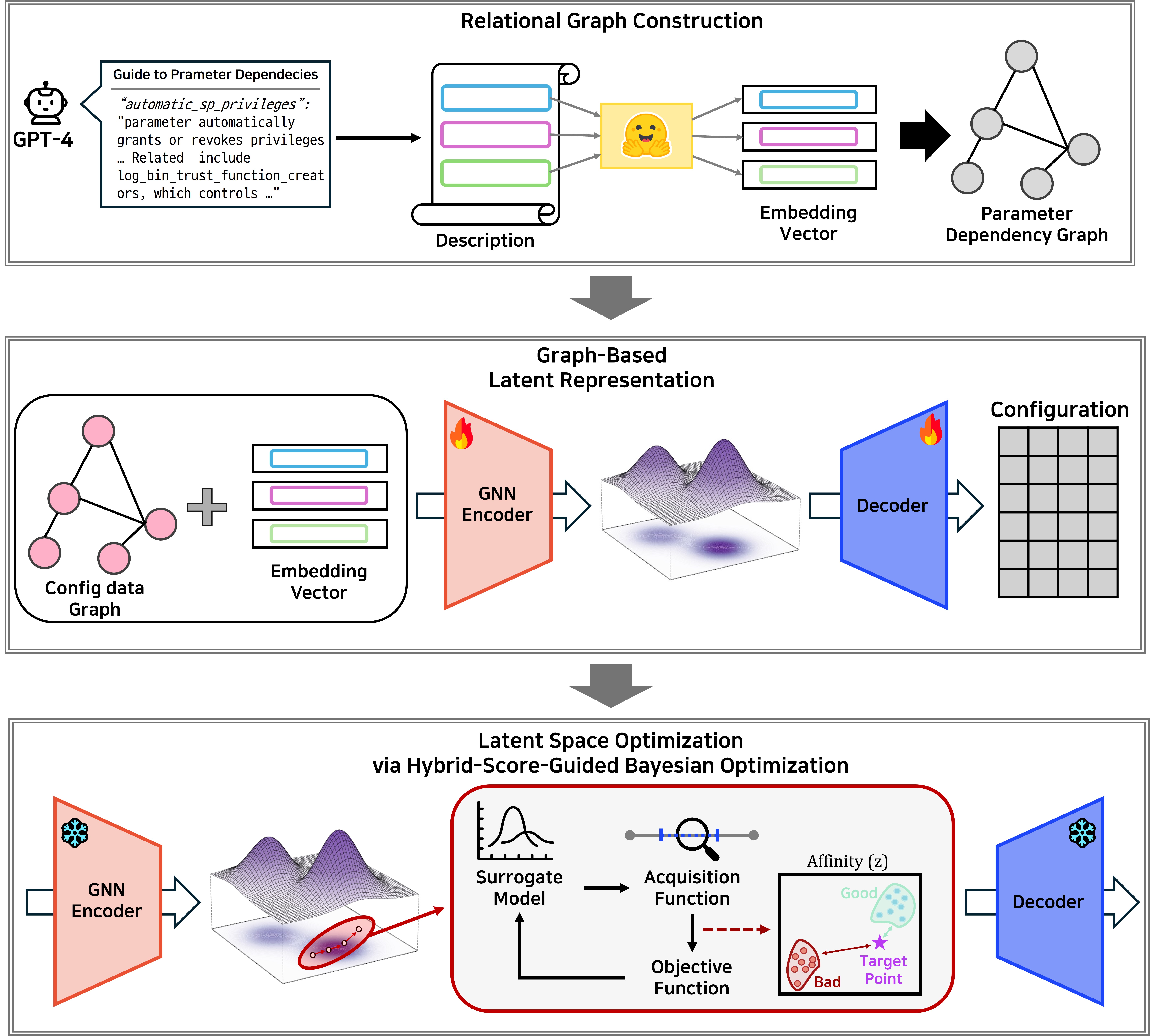}
    \caption{Overview Architecture of RelTune}
    \label{fig:overview}
\end{figure*}

\section{Method}
\subsection{Overview}


We propose RelTune, a novel framework that efficiently explores the entire parameter space by representing semantic relationships among parameters as a graph, encoding them into a Graph Neural Network (GNN)-based \cite{GNN} latent representation, and performing Hybrid Score–Guided Bayesian Optimization (Hybrid BO). The overall architecture of RelTune is illustrated in Figure \ref{fig:overview}, which consists of three main components.

\textbf{Step 1: Relational Graph Construction.} In the first stage, we construct a parameter relationship graph that captures semantic dependencies among DBMS configuration parameters. Specifically, we first use a Large Language Model (LLM; GPT-4 \cite{GPT}) to generate textual descriptions that capture the relationships among parameters.
These descriptions are then transformed into embedding vectors through the LLM API, and pairwise cosine similarity between the embeddings is computed to form edges in the relational graph. 

\textbf{Step 2: Graph-Based Latent Representation.} In the second stage, we inject configuration values into the nodes of the relational graph to generate multiple samples and train a GNN. The GNN learns to encode both parameter dependencies and performance metrics into a latent representation. Additionally, a decoder is trained to reconstruct the original configuration from the latent representation, ensuring that the learned embedding preserves the structural information of the original parameter space.

\textbf{Step 3: Latent Space Optimization via Hybrid-Score-Guided Bayesian Optimization.} In the third stage, optimization is performed over the latent representations generated in Step 2. Instead of applying Vanilla Bayesian Optimization (VBO), we adopt a Hybrid-Score-Guided BO (HBO) that integrates an Affinity Score, which reflects the relational proximity between candidate points and high-performing configurations. This approach enables more efficient exploration by focusing the search on promising regions of the latent space associated with superior performance.

\subsection{Relational Graph Construction}
\subsubsection{\bfseries Parameter Embedding}
To extract semantic relationships among parameters across various DBMSs, we leverage extensive domain knowledge sources such as official DBMS manuals and community forum discussions through a Large Language Model (LLM), GPT-4 \cite{GPT}. We prompt the model with queries of the form: 
\begin{quote}
\texttt{For a \textcolor{blue!70!black}{``given DBMS (specify the version)''}, explain the functionality of each parameter and list
related parameters that affect its behavior.}
\end{quote}





\noindent This prompt is designed to elicit functional descriptions and inter-parameter dependencies directly from domain texts, allowing the LLM to capture semantically grounded relationships among configuration parameters. Including the specific DBMS version in the prompt helps ensure that the extracted descriptions and parameter relationships reflect version-specific behaviors.
Although most parameter dependencies are generally consistent across versions, version-level context provides more precise and up-to-date information, thereby improving the reliability of the extracted knowledge.

This process automatically generates textual descriptions for each parameter.
For example, when queried about the parameter \textit{\texttt{automatic\_sp\_privileges}}, the model produces an output in the following format:

\begin{lstlisting}[caption={Example LLM Output}, label={lst:llm-prompt}]
"automatic_sp_privileges": "parameter automatically grants or revokes privileges for stored procedures when they are created or dropped. It simplifies stored procedure privilege management. Related  include log_bin_trust_function_creators, which controls whether function creators must have SUPER privilege."
\end{lstlisting}


The generated descriptions serve as sentence-level data that encapsulate the semantic characteristics and relational information of each parameter. To embed these descriptions, we employ the paraphrase-MiniLM-L6-v2  \cite{minilm} model from the Sentence-Transformers library. This model, fine-tuned on paraphrase datasets, is designed to capture contextual semantic similarity rather than mere lexical overlap \cite{minilm_paper}, making it well-suited for measuring functional relationships among DBMS parameters. In particular, the Sentence Transformer architecture is optimized such that cosine similarity directly reflects the semantic distance between embeddings, which aligns naturally with our graph construction approach that defines edges based on similarity thresholds. Moreover, the lightweight MiniLM structure ensures high computational efficiency, enabling scalable embedding even for large sets of parameters.

\subsubsection{\bfseries Graph Construction}
For the embedding vectors generated in the previous step, let \(\mathbf{e}_i\) and \(\mathbf{e}_j\) denote the embedding representations of parameters \(i\) and \(j\), respectively. The semantic similarity between two parameters is defined using cosine similarity, as follows:

\begin{equation}
S_{ij} = \frac{\mathbf{e}_i \cdot \mathbf{e}_j}{\lVert \mathbf{e}_i \rVert \, \lVert \mathbf{e}_j \rVert}
\end{equation}

This similarity represents the probabilistic proximity of two parameters belonging to the same functional context, and it is used to define the edges of the relational graph. The relational graph $G = (V,E)$ consists of a set of parameters $V$ and their relationship $E$. Each node corresponds to a parameter, and an edge between parameters $i$ and $j$ is established based on their cosine similarity $S_{ij}$. The binary adjacency matrix $A \in \mathbb{R}^{n \times n}$ is defined as:

\begin{equation}
A_{ij} =
\begin{cases}
1, & \text{if } S_{ij} \ge \tau,\\
0, & \text{otherwise,}
\end{cases}
\label{eq:adjacency}
\end{equation}

where $\tau$ is the similarity threshold controlling edge creation. If the similarity exceeds $\tau$, an edge is created between the corresponding nodes; 
otherwise, no edge is added.

In this study, the cosine similarity threshold \(\mathbf{\tau}\) is set to 0.75, which was empirically found to provide the best balance between graph density and modularity (\ref{graph quality}). The graph constructed with this threshold prevents excessive connectivity while encouraging semantically related parameters to form clusters within the same subsystem (e.g., InnoDB, Query Optimizer). Consequently, the resulting relational graph structurally encodes the semantic dependencies and functional correlations among parameters, providing a solid foundation for the GNN-based encoder to effectively model parameter relationships.

\subsection{Graph Based Latent Representation}
\subsubsection{\bfseries Relational Graph Representation Learning}
To compress the high-dimensional relational graph of parameters into a low-dimensional latent space, we employ a GNN–based encoder. To reconstruct the original high-dimensional configuration from this latent representation, an autoencoder-style decoder is additionally trained. (\ref{performance_aware})

Based on the relational graph constructed in the previous stage, we inject the parameter values of each configuration sample into the corresponding nodes. The embedding vectors of all nodes are then concatenated to jointly reflect both the numerical and semantic characteristics of the parameters.
Through this design, each node simultaneously encodes a parameter’s quantitative and semantic properties, while all configuration samples share the same edge structure $E$ but differ in their feature distributions.
As a result, each configuration sample can be represented as an instance-level graph.

To learn latent representations from the constructed graphs, we employ an encoder based on the Graph Attention Network (GAT) \cite{gat} architecture.
GAT enables each node to selectively aggregate information from its neighbors using attention weights, rather than a simple average, allowing the model to learn the relative importance of parameter interactions. The node representation at layer $l$ is defined as follows:

\begin{equation}
\mathbf{e}_i^{(l+1)}
  = \sigma\!\left(
    \sum_{j \in \mathcal{N}(i)}
      \alpha_{ij}^{(l)}\,
      \mathbf{W}^{(l)} \mathbf{e}_j^{(l)}
  \right)
\end{equation}

The ${W}^{(l)}$ denotes a trainable linear transformation, and  \( \sigma(\cdot) \) represents a nonlinear activation function (e.g., ELU). The attention coefficient  $\alpha_{ij}^{(l)}$, which indicates the importance of node $j$ information when being aggregated into node $i$, is computed using a softmax-based normalization as follows:

\begin{equation}
\alpha_{ij}^{(l)}
= \operatorname{softmax}_{j}\!\left(
    \operatorname{LeakyReLU}\!\Big(
      \mathbf{a}^{\top}
      \big[
        \mathbf{W}^{(l)}\mathbf{e}_i^{(l)} \concat
        \mathbf{W}^{(l)}\mathbf{e}_j^{(l)}
      \big]
    \Big)
  \right).
\end{equation}

This attention mechanism enables the model to learn anisotropic relationships among parameters, allowing it to emphasize semantically important interactions rather than relying solely on structural connectivity.
In the final layer, all node embeddings are aggregated via average pooling to produce a latent vector $z$ that represents the entire graph.

\begin{equation}
\mathbf{z} = \frac{1}{|\mathcal{V}|} \sum_{i \in \mathcal{V}} \mathbf{e}_i^{(L)}
\end{equation}

This latent vector serves as a compact representation that summarizes the overall performance patterns and relational dependencies among database parameters.
It is subsequently used as the search space for the optimization phase.

\subsubsection{\bfseries Performance Aware Latent Reconstruction} \label{performance_aware}
To reconstruct the latent vector $z$ into its original configuration form, we train an autoencoder-style decoder \cite{AE}. The decoder trains a mapping from the latent space back to the original high-dimensional configuration, denoted as \(\hat{x} = \mathcal{D}(z)\), thereby encouraging the latent representation to preserve the continuity and semantic consistency across configurations. In addition, to ensure that the latent space captures not only configuration structure but also performance variations, we train a metric prediction head $e_\psi$ in parallel to predict throughput and latency directly from the latent vector. This auxiliary objective injects performance-aware information into the latent representation,
enabling it to reconstruct not only the configuration itself but also the underlying structural relationships between parameter values and performance outcomes. The decoder and the metric prediction head are jointly trained to minimize the loss function.

\begin{equation}
\mathcal{L}
= \lambda_{recon} \lVert x - \hat{x} \rVert_2^2
+ \lambda_{metric} \lVert y - \hat{y} \rVert_2^2
\end{equation}

Here, the reconstructed configuration and the predicted performance are denoted as  \( \hat{x}\) and \( \hat{y} = e_{\psi}(z) \), respectively, while  \(\lambda_{recon}\) and \(\lambda_{metric}\) are hyper-parameters that control the relative importance of the two terms. The reconstruction loss encourages the latent space to preserve the structural properties of the original configuration space, whereas the metric prediction loss aligns the latent space such that configurations with similar performance are positioned close to each other.

\subsection{Latent Space Optimization via Hybrid-Score-Guided Bayesian Optimization}
Conventional BO determines the next search region based solely on the predictions of the surrogate model within the latent space. However, this approach often requires a large number of exploration steps to achieve reliable predictions. In high-dimensional search space, such as database parameters, BO becomes inefficient and is prone to getting trapped in local minima. To overcome these limitations, we propose a Hybrid-Score-Guided Bayesian Optimization (HBO) framework that combines the surrogate model’s predicted performance score with an affinity score, which measures the proximity between a candidate latent vector and the set of previously identified high-performing latent vectors. Formally, for each latent vector $z$, the Hybrid Score is defined as follows:

\begin{equation}
\text{HybridScore}(z)
= f_{\text{metric}}(z)
+ \gamma \cdot f_{\text{affinity}}(z)
\end{equation}

Here, $f_{\text{metric}}$ denotes the performance score computed from the predicted throughput and latency obtained via the metric prediction head described in Section \ref{performance_aware} (Performance Aware Latent Reconstruction). Since higher throughput and lower latency indicate better performance, the score is defined as follows:

\begin{equation}
f_{\text{metric}}(z)
= f_{\text{Throughput}}(z) - \alpha \cdot f_{\text{Latency}}(z),
\end{equation}

where, $\alpha$ is a hyperparameter that controls the importance of latency in the performance score computation.
In addition, $f_{\text{affinity}}(z)$ measures how close a candidate latent vector $z$ is to the set of previously identified high performing latent vectors. Rather than forcing convergence toward a single mean point,as in conventional BO, this term encourages exploration within regions densely populated by good samples, enabling more fiverse and performance-aware search. Specifically, let $Z_{good}$ denote the set of latent vectors whose performance exceeds a predefined threshold. The affinity score is the computed as follows:

\begin{equation}
f_{\text{affinity}}(z) = \frac{1}{|Z_\text{{good}}|} \sum_{z_\text{k} \in Z_\text{{good}}} \exp\!\left(-\frac{\lVert z - z_k \rVert^2}{2\sigma^2}\right),
\end{equation}

where the distance between $z$ and each \( z_k \in \mathcal{Z}_{\text{good}} \) is converted into a similarity value using an RBF (Radial Basis Function) kernel, and the average is taken over all high-performing samples. Consequently, a candidate vector $z$ located closer to high-performance regions yields a higher affinity score. 

The hybrid score achieves two complementary effects. First, $f_{\text{metric}}(z)$ enables exploitative exploration within regions of the latent space that are predicted to yield high performance. Second, $f_{\text{affinity}}(z)$ encourages exploration toward regions of the latent space that are densely populated by previously high-performing configurations, thereby facilitating the discovery of new optimal configurations. (Algorithm \ref{alg:hbo}, L6-L17)

Based on the defined hybrid score, BO is performed to identify the optimal latent vector within the latent space. Within each BO loop, the hybrid score serves as the objective function for training the surrogate model. A GP–based surrogate is iteratively updated to approximate the hybrid score, and the Expected Improvement (EI) acquisition function is employed to select the next exploration point with the highest potential for improvement. The EI function measured the expected gain over the current best hybrid score and is defined as follows:



\begin{multline}
\mathrm{EI}(z)
= \mathbb{E}\!\left[\max\!\bigl(0,\, f(z)-f(z^+)\bigr)\right] \\
= \bigl(\mu(z)-f(z^+)\bigr)\,\Phi\!\bigl(\gamma(z)\bigr)
  + \sigma(z)\,\phi\!\bigl(\gamma(z)\bigr),
\end{multline}

where $\mu(z)$ and $\sigma(z)$ denote the predictive mean and standard deviation of the GP, $f(z^+)$ represents the current best hybrid score, and \( \Phi(\cdot) \) and \( \phi(\cdot) \) are the cumulative and probability density functions of the standard normal distribution, respectively.

Since direct performance evaluation of each optimized $z$ is computationally expensive,
the actual performance is predicted using the pretrained GAT-based metric prediction head instead of conducting costly benchmark executions. (Algorithm \ref{alg:hbo}, L9-10)
 

\begin{algorithm}[H]
\caption{Hybrid-Score-Guided Bayesian Optimization (HBO)}
\label{alg:hbo}
\begin{algorithmic}[1] 
\Require 
    Pre-trained GNN model $\mathcal{M}$ (with $f_\text{{enc}}, g_\text{{dec}}$, and $e_\psi$) \\
    Initial dataset $\mathcal{D}$ (with graph $\text{G}$, performance $\text{P}$; Hyperparameters $\alpha, \gamma, \sigma$; Iterations $\text{T}$
\Ensure 
    Optimized configuration $c^*$

\State $Z_\text{{all}} \gets \{f_\text{{enc}}(G_i) \mid G_i \in \mathcal{D}\}$
\State $Z_\text{{good}} \gets \{z_\text{i} \in Z_\text{{all}} \mid \text{is\_good}(\text{performance}_i)\}$
\State Define search space $\mathcal{S}$ from the bounds of $Z_\text{{all}}$
\State Initialize Gaussian Process (GP) model $\mathcal{GP}$

\For{$t = 1$ to $T$}
    \State $z_\text{t} \gets \arg\max_{z \in \mathcal{S}} \text{EI}(z \mid \mathcal{GP})$
    
    \State Calculate Hybrid Score for the proposed point $z_t$
    \State $f_\text{{TPS}}(z_\text{t}), f_\text{{Latency}}(z_\text{t}) \gets \text{$e_\psi$}(z_\text{t})$
    \State $f_\text{{metric}}(z_\text{t}) \gets f_\text{{TPS}}(z_\text{t}) - \alpha \cdot f_\text{{Latency}}(z_\text{t})$
    \State $f_\text{{affinity}}(z_\text{t}) \gets \frac{1}{|Z_\text{{good}}|} \sum_{z' \in Z_\text{{good}}} \exp\left(-\frac{\|z_t - z'\|^2_2}{2\sigma^2}\right)$
    \State $\text{HybridScore}(z_\text{t}) \gets f_\text{{metric}}(z_\text{t}) + \gamma \cdot f_\text{{affinity}}(z_\text{t})$
    
    \State Update $\mathcal{GP}$ with the observation $(z_\text{t}, \text{HybridScore}(z_\text{t}))$
\EndFor

\State $z^* \gets \text{the latent vector with the best observed HybridScore}$
\State $c^* \gets g_\text{{dec}}(z^*)$
\State \Return $c^*$
\end{algorithmic}
\end{algorithm}

\section{Experiment}
\subsection{Experiment Setup} \label{Experiment Setup}

\noindent{\textbf{Hardware.}} All MySQL and PostgreSQL performance measurements were performed on a server running the Ubuntu 20.04.6 operating system with an Intel® Core™ i7-11700 @ 2.52GHZ, 32 GB of RAM, and a 256 GB disk. \\

\noindent\textbf{Tuning Setting.} 
We conducted experiments using \textit{MySQL v5.7.37}~\cite{mysql} and \textit{PostgreSQL v14.19}~\cite{Postgres}. 
For all experiments, the models were trained on a dataset consisting of 5,000 configuration and performance samples. 
Each workload experiment was repeated five times with different random seeds, and the reported results represent the average of these runs. 
For baselines that optimize only the top-ranked parameters, the number of selected parameters $k$ was set to 5, 10, and 15, respectively, and the best-performing result among them was used for comparison. \\

\noindent\textbf{Baselines.} 
We compared our method with the following representative tuning frameworks: 
\textbf{OtterTune}~\cite{OtterTune}, \textbf{RGPE}~\cite{Facilitating}, and \textbf{CSAT}~\cite{csat}, which employ BO over a subset of top parameters; 
\textbf{GPTuner}~\cite{gptuner}, which leverages large language models to extract and convert domain knowledge for BO-based tuning; 
and \textbf{CDBTune}~\cite{CDBTune}, which performs RL-based tuning. 
All baselines were executed under identical hardware and workload conditions for a fair comparison. \\

\noindent\textbf{Workloads.} 
We evaluated the proposed method on four workloads for MySQL and two workloads for PostgreSQL. 
For MySQL, we adopted multiple combinations of \textbf{YCSB (Yahoo! Cloud Serving Benchmark)}~\cite{ycsb} workloads, which define diverse ratios of database operations such as \textit{load}, \textit{insert}, \textit{read}, \textit{update}, and \textit{scan}. 
The workload configurations are summarized in Table~\ref{tab:ycsb_workload}. 
For PostgreSQL, we used two representative workloads: the OLTP workload \textbf{TPC-C} \cite{tpcc} and the OLAP workload \textbf{TPC-H} \cite{tpch}. 
All workloads were executed using \textbf{BenchBase}~\cite{benchbase, benchbase_paper}, which reports throughput as \textit{transactions per second (TPS)} by treating each execution unit, including individual queries in TPC-H, as a single transaction. 
To ensure consistency, we uniformly report throughput in TPS units across all workloads, together with corresponding latency results. \\

\noindent{\textbf{Implementation Details.}} Our model projected to 32 dimensions for MySQL and PostgreSQL. We set the same 300 iterations for all optimization processes in our experiments. In the performance function $f_{\text{metric}}$, we set $\alpha = 0.5$ to reflect the trade-off between throughput and latency, ensuring a balanced optimization objective between the two metrics. 
In the Hybrid Score, the parameter $\gamma = 1.0$ was chosen to perform balanced exploration by equally weighting the metric prediction and the affinity score. These hyperparameters are user-adjustable, allowing flexible trade-offs between throughput, latency, and exploration intensity depending on the workload characteristics.

\begin{table}[t]
    \centering
    \caption{MySQL Workload Information}
    \label{tab:ycsb_workload}
    \small
    \setlength{\tabcolsep}{2pt} 
    \setlength{\extrarowheight}{0pt} 
    \begin{tabular}{cccccccc}
        \hline
        \textbf{MySQL} & \textbf{} & \textbf{} & \multicolumn{1}{l}{} & \multicolumn{1}{l}{} & & \multicolumn{1}{l}{} & \multicolumn{1}{l}{} \\ \hline
        \multicolumn{1}{c|}{\begin{tabular}[c]{@{}c@{}}Workload\\Index\end{tabular}} 
        & \multicolumn{1}{c|}{\begin{tabular}[c]{@{}c@{}}Scale\\Factor\end{tabular}} 
        & \multicolumn{1}{c|}{\begin{tabular}[c]{@{}c@{}}Data\\Size\end{tabular}} 
        & \multicolumn{1}{l|}{Read} & \multicolumn{1}{l|}{Insert} 
        & \multicolumn{1}{c|}{Scan} & \multicolumn{1}{l|}{Update} 
        & \multicolumn{1}{l}{\begin{tabular}[c]{@{}l@{}}Read\\Modify\\Write\end{tabular}} \\ \hline
        \multicolumn{1}{c|}{A} & \multicolumn{1}{c|}{\multirow{4}{*}{12000}} & \multicolumn{1}{c|}{\multirow{4}{*}{15GB}} 
        & \multicolumn{1}{c|}{\textbf{50\%}} & \multicolumn{1}{c|}{-} 
        & \multicolumn{1}{c|}{-} & \multicolumn{1}{c|}{\textbf{50\%}} & - \\
        \multicolumn{1}{c|}{B} & \multicolumn{1}{c|}{} & \multicolumn{1}{c|}{} & \multicolumn{1}{c|}{\textbf{95\%}} & \multicolumn{1}{c|}{-} 
        & \multicolumn{1}{c|}{-} & \multicolumn{1}{c|}{\textbf{5\%}} & - \\
        \multicolumn{1}{c|}{E} & \multicolumn{1}{c|}{} & \multicolumn{1}{c|}{} & \multicolumn{1}{c|}{-} & \multicolumn{1}{c|}{\textbf{5\%}} 
        & \multicolumn{1}{c|}{\textbf{95\%}} & \multicolumn{1}{c|}{-} & - \\
        \multicolumn{1}{c|}{F} & \multicolumn{1}{c|}{} & \multicolumn{1}{c|}{} & \multicolumn{1}{c|}{\textbf{50\%}} & \multicolumn{1}{c|}{-} 
        & \multicolumn{1}{c|}{-} & \multicolumn{1}{c|}{-} & \textbf{50\%} \\ \hline
    \end{tabular}
\end{table}

\begin{figure}[t]  
    \begin{minipage}{\columnwidth}  
        \centering
        \includegraphics[width=1.0\columnwidth,height=0.5\textheight]{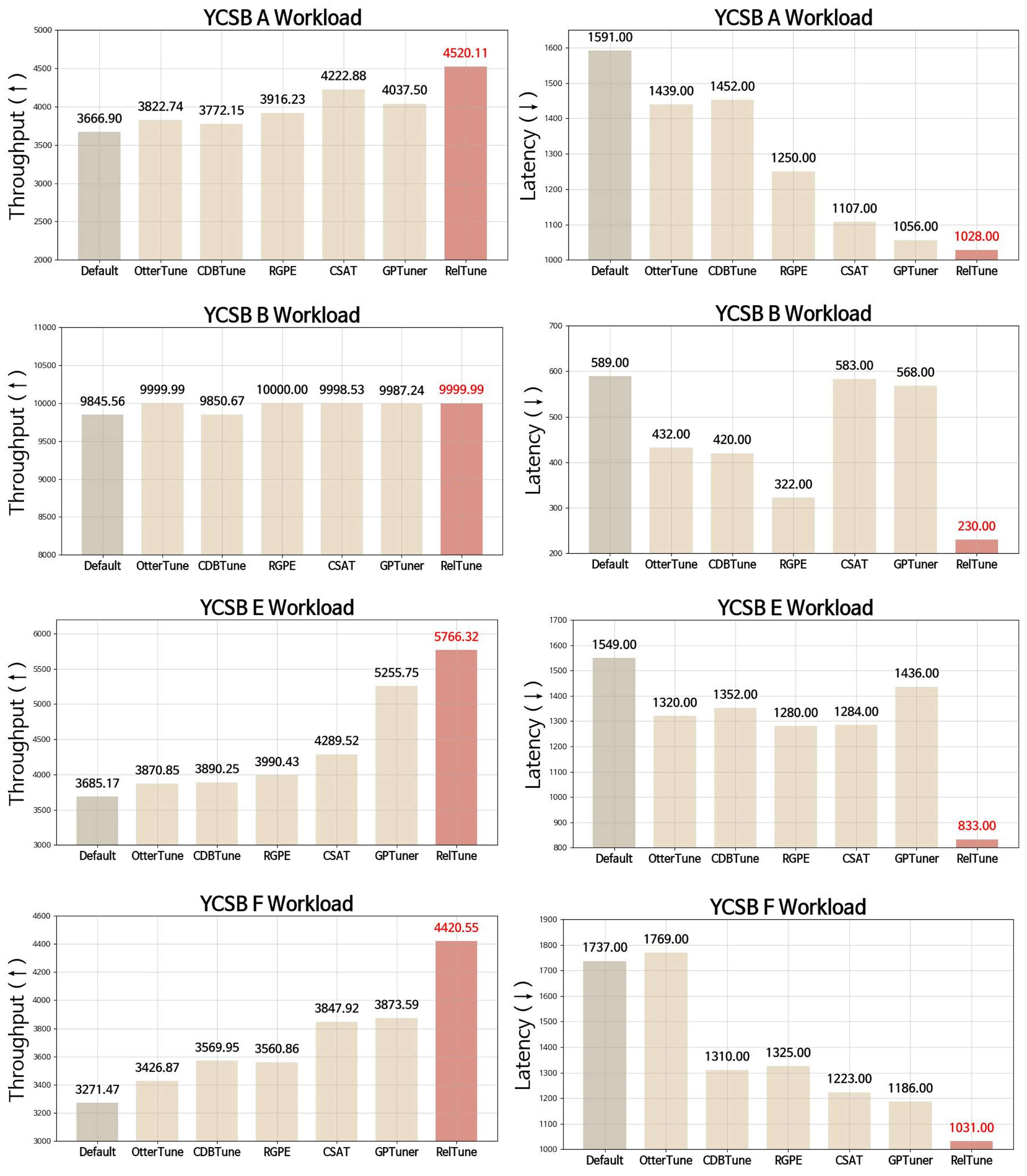}  
        \caption{Performance improvement for MySQL YCSB A,B,E and F.}
        \label{fig:mysql_output}
    \end{minipage}
\end{figure}

\begin{figure}[t]  
    \begin{minipage}{\columnwidth}  
        \centering
        \includegraphics[width=1.0\columnwidth]{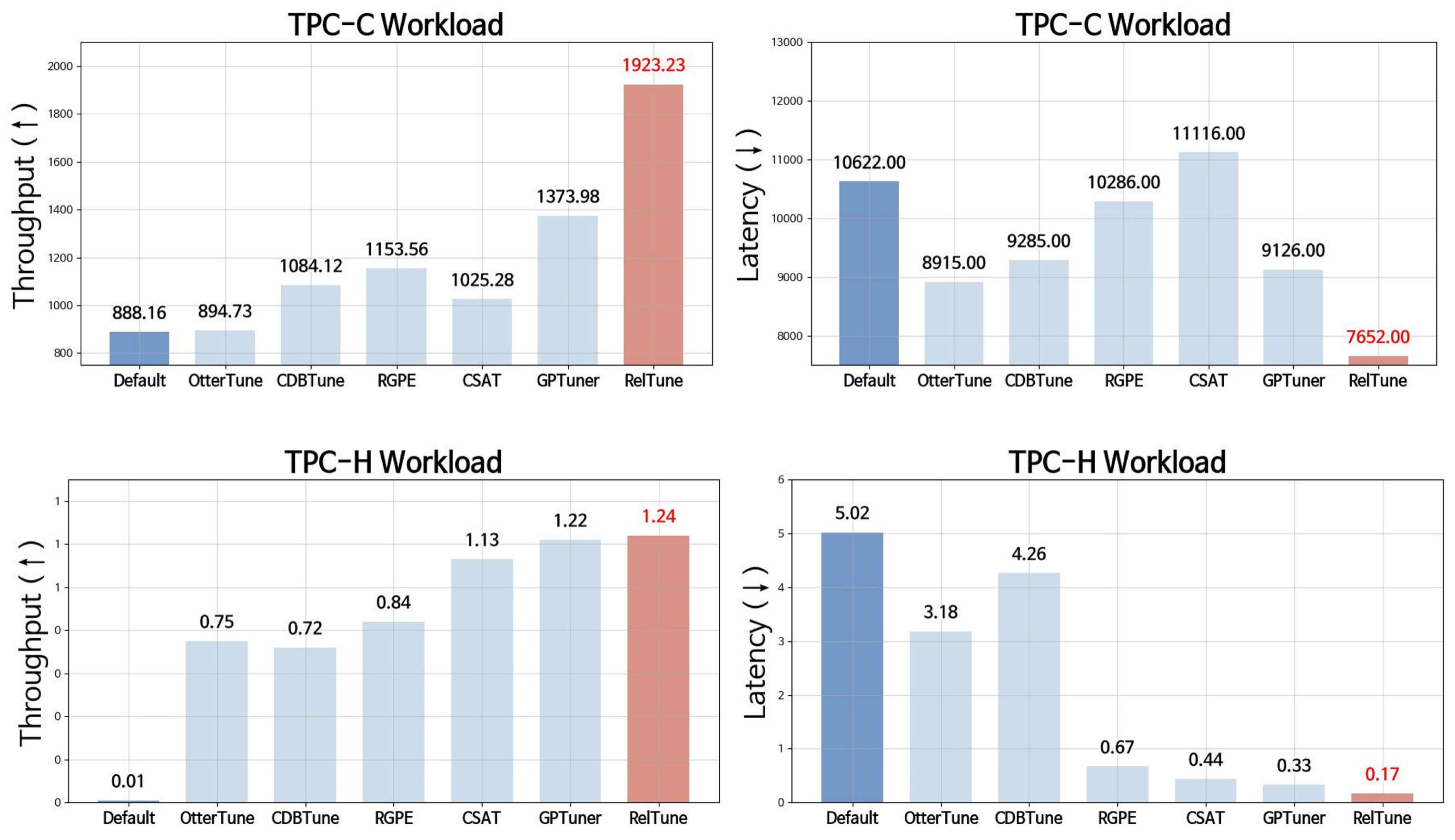}  
        \caption{Performance improvement for PostgreSQL TPC-C and TPC-H.}
        \label{fig:postgres_output}
    \end{minipage}
\end{figure}



\subsection{Performance Comparison}
Figure \ref{fig:mysql_output} and \ref{fig:postgres_output} present the throughput and latency results across different workloads for both MySQL and PostgreSQL. For MySQL, RelTune consistently outperforms all baseline methods across four YCSB workloads (YCSB A, B, E, and F). In particular, under YCSB E, RelTune achieves the largest performance gain, improving throughput by 56.5\% compared to the default configuration. YCSB B is a read-heavy workload with a very low proportion of write or update operations. Due to this characteristic, the influence of parameters related to internal buffering or log writing is relatively limited, resulting in a small improvement in throughput and other baselines. Nevertheless, RelTune achieves a substantial reduction in latency, with a 60.9\% improvement compared to the default performance.

Across both the OLTP workload (TPC-C) and the OLAP workload (TPC-H), RelTune shows superior performance in terms of both throughput and latency. To maintain a consistent evaluation metric between the MySQL and PostgreSQL experiments, we adopted a unified throughput-based evaluation framework for both DBMSs.
This consistency was necessary because evaluating TPC-H using only latency or QPS would make it difficult to directly compare the relative performance improvements with other workloads, such as YCSB.



As a result, RelTune exhibits higher performance stability and superior optimization capability compared to both traditional BO-based or RL-based tuning approaches and recent baseline models that optimize only a subset of parameters. Moreover, RelTune consistently delivers performance improvements across both OLTP and OLAP environments, demonstrating its robustness and general applicability.



\subsection{Graph Quality Analysis} \label{graph quality}
We constructed the parameter relationship graph by connecting edges only between parameter nodes whose embedding cosine similarity exceeded 0.75. To validate the effectiveness of the threshold, we conducted a comprehensive evaluation of the structural and semantic quality of the resulting graph from multiple perspectives.

First, we evaluated the structural cohesiveness of the graph by computing its modularity ($Q$). Modularity measures how strongly the nodes are grouped into communities by comparing the proportion of intra-community edges in the actual graph to that expected in a random graph with the same number of nodes and edges. A higher $Q$ value indicates that the graph exhibits a well-defined community structure, implying that edges are not randomly connected but instead form functionally cohesive groups. The modularity $Q$ is defined as follows:

\begin{equation}
Q = \frac{1}{2m} \sum_{i,j} \left[A_{ij} - \frac{k_i k_j}{2m}\right] \delta(c_i, c_j),
\end{equation}

Here, $A_{ij}$ denoted the element of the adjacency matrix, $k_i$ and $k_j$ represent the degrees of nodes $i$ and $j$, respectively, $m$ is the total number of edges, and $\delta(c_i, c_j)$ is an indicator function that equals 1 if nodes $i$ and $j$ belong to the same community, and 0 otherwise.

Additionally, we we evaluated the semantic consistency of the graph by applying the Louvain community detection algorithm. To measure the alignment between the communities identified by the Louvain algorithm and the predefined parameter subsystems, we employed two clustering evaluation metrics: Normalized Mutual Information (NMI) and Adjusted Rand Index (ARI). Here, each subsystem was defined by grouping parameters according to their prefixes (e.g., innodb\_..., optimizer\_...). NMI measures the overall consistency between the detected community structure and the predefined subsystems, while ARI quantifies the pairwise labeling agreement adjusted for random chance. Higher values of both metrics indicate stronger semantic alignment between the graph’s community structure and the actual subsystem organization. 

However, since the subsystem prefixes of parameters do not always correspond to their functional relationships, we jointly considered modularity, representing structural cohesiveness, and NMI, ARI, reflecting semantic consistency, to comprehensively validate the quality of the constructed graph.

\newcommand{\smileflag}{\llap{\hspace{0.15em}\smiley\,}}

\begin{table}[t]
\centering
\caption{Graph quality metrics of the MySQL parameter graph.}
\label{tab:my-graph}
\small                      
\setlength{\tabcolsep}{8pt}     
\renewcommand{\arraystretch}{1.2}
\begin{tabular}{c|c|c|c}
\toprule
\textbf{Threshold} & \textbf{Modularity} & \textbf{NMI} & \textbf{ARI} \\
\midrule[\heavyrulewidth]
0.65 & 0.45 & 0.45 & 0.21 \\
0.70 & 0.61 & 0.45 & 0.11 \\
\rowcolor{blue!10}
\smileflag\textbf{0.75} & \textbf{0.77} & \textbf{0.46} & \textbf{0.12} \\
0.80 & 0.87 & 0.46 & 0.03 \\
0.85 & 0.92 & 0.45 & 0.01 \\
\bottomrule
\end{tabular}
\end{table}

\begin{table}[t]
\centering
\caption{Graph quality metrics of the PostgreSQL parameter graph.}
\label{tab:post-graph}
\small                     
\setlength{\tabcolsep}{8pt}     
\renewcommand{\arraystretch}{1.2}
\begin{tabular}{c|c|c|c}
\toprule
\textbf{Threshold} & \textbf{Modularity} & \textbf{NMI} & \textbf{ARI} \\
\midrule[\heavyrulewidth]
0.65 & 0.72 & 0.58 & 0.32 \\
0.70 & 0.79 & 0.60 & 0.25 \\
\rowcolor{blue!10}
\smileflag \textbf{0.75} & \textbf{0.89} & \textbf{0.60} & \textbf{0.17} \\
0.80 & 0.91 & 0.56 & 0.02 \\
0.85 & 0.81 & 0.55 & 0.01 \\
\bottomrule
\end{tabular}
\end{table}

Table \ref{tab:my-graph} and \ref{tab:post-graph} summarize the evaluation results for the parameter relationship graphs of MySQL and PostgreSQL, where the cosine similarity threshold was varied from 0.65 to 0.85. A comprehensive analysis of the three evaluation metrics (Modularity, NMI and ARI) reveals a common trend across MySQL and PostgreSQL: as the threshold increases, Modularity improves initially, but ARI begins to drop sharply beyond a certain point.

For MySQL, the modularity reached its peak value (0.92) when the threshold was set to 0.85. However, at this point, the ARI sharply dropped to 0.01, indicating severe over-segmentation of communities. This suggests that the graph became excessively sparse, leading to the loss of essential inter-subsystem connections.

In contrast, PostgreSQL exhibited more moderate variations in the evaluation metrics as the threshold changed,
which can be attributed to its more homogeneous correlation structure among parameters compared to MySQL. Specifically, when the threshold was set to 0.80, the modularity increased to 0.91, but the ARI sharply declined to 0.02, indicating a degradation in the semantic alignment of the graph.




In summary, Taken together, these results confirm that a threshold of 0.75 provides the optimal balance between structural cohesiveness (Modularity) and semantic consistency (ARI) for both DBMSs. For MySQL, the modularity remained sufficiently high at 0.77, while the ARI stayed stable, and for PostgreSQL, both the modularity (0.89) and ARI (0.17) maintained steady levels, preserving subsystem-level semantics without excessive fragmentation.
Accordingly, this study adopts 0.75 as the common threshold, as it best captures the trade-off between semantic relatedness and structural stability in constructing parameter relationship graphs for both DBMSs.

\subsection{Validation of the Affinity Score}
To verify whether the proposed affinity score reliably reflects the relationship between the latent representations and the actual performance, we analyzed and evaluated the correlation between the affinity score and the normalized throughput and latency. Specifically, we selected the top-30 and bottom-30 configurations based on performance (throughput and latency) and converted them into their corresponding latent vectors. Using these two sets, we computed the affinity score of each sample by measuring its distance to the latent vectors of previously well-performing configurations. This evaluation allows us to verify whether higher affinity scores are indeed associated with better throughput and lower latency in practice.

For quantitative evaluation, we employed AUROC (Area Under the ROC Curve) and AUPRC (Area Under the Precision–Recall Curve), and additionally performed visual analysis to support the quantitative findings. AUROC represents the probability that the model correctly distinguishes between high-performing and low-performing latent vectors, while AUPRC measures the balance between precision and recall for identifying high-performing latent vectors. Both metrics range from 0 to 1, with higher values indicating a stronger ability to discriminate well-performing configurations. In this study, these metrics were used to evaluate whether the proposed Affinity Score exhibits a strong correlation with actual performance. The dashed line in the figure represents the binned average trend, which shows the average performance variation within each affinity interval.

\begin{figure}[t]
    \centering
    \includegraphics[width=\columnwidth]{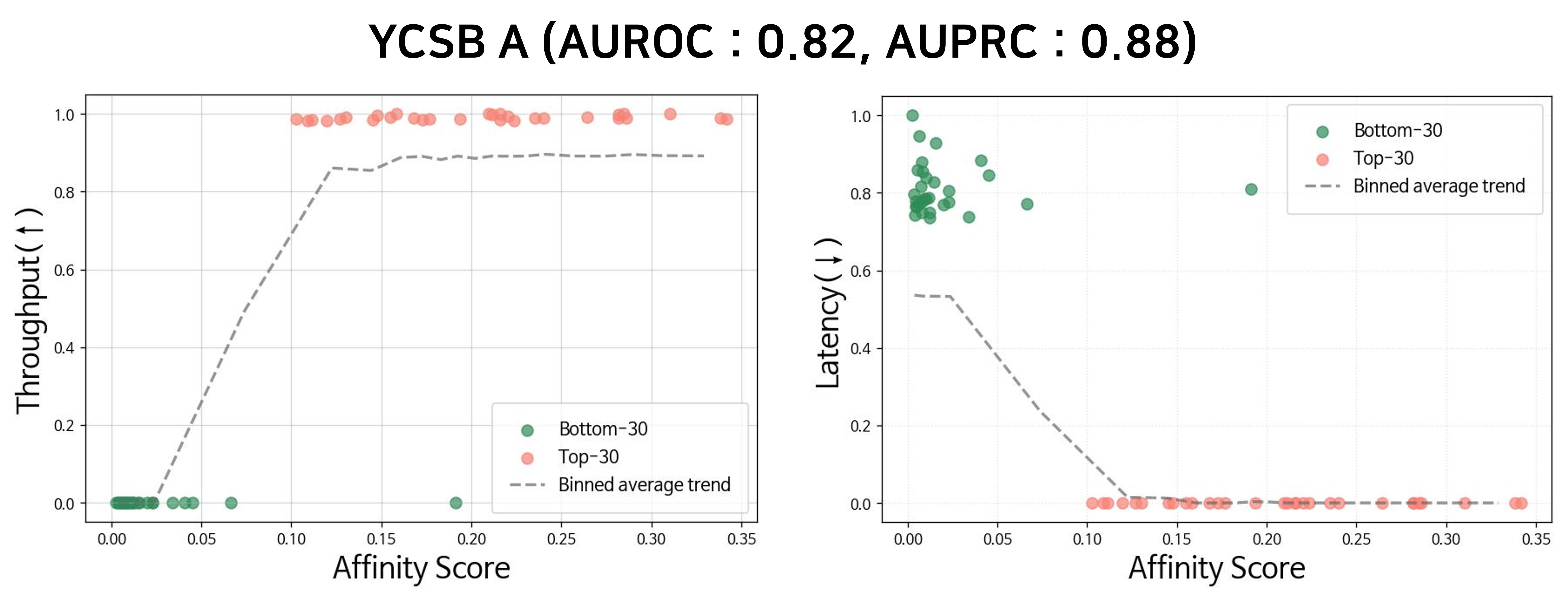}
    \vspace{0.5em}
    \includegraphics[width=\columnwidth]{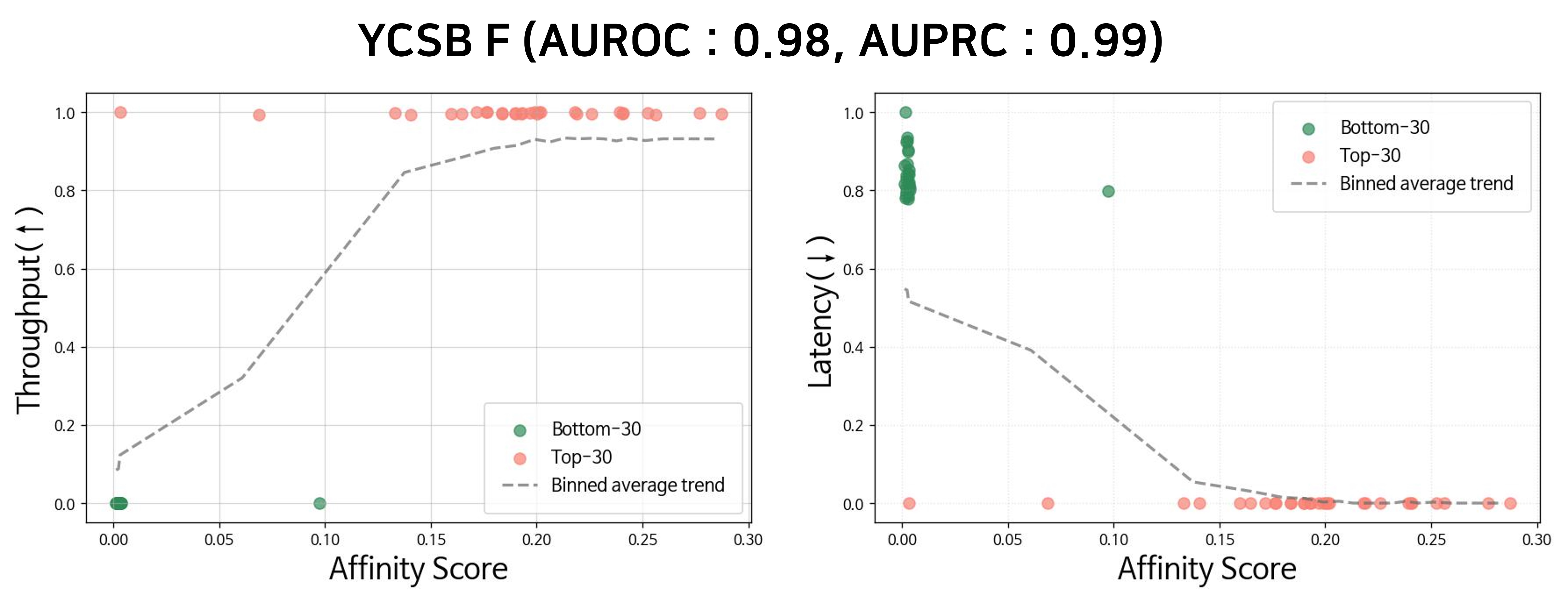}
    \caption{Correlation between Affinity Score and Performance Metrics for MySQL.}
    \label{fig:affinity_ex}
\end{figure}

Figure \ref{fig:affinity_ex} presents the results for MySQL on the YCSB A and YCSB F workloads. The upper plot corresponds to the YCSB A workload, while the lower plot shows the result for YCSB F.

\begin{figure}[t]  
    \begin{minipage}{\columnwidth}  
        \centering
        \includegraphics[width=1.0\columnwidth]{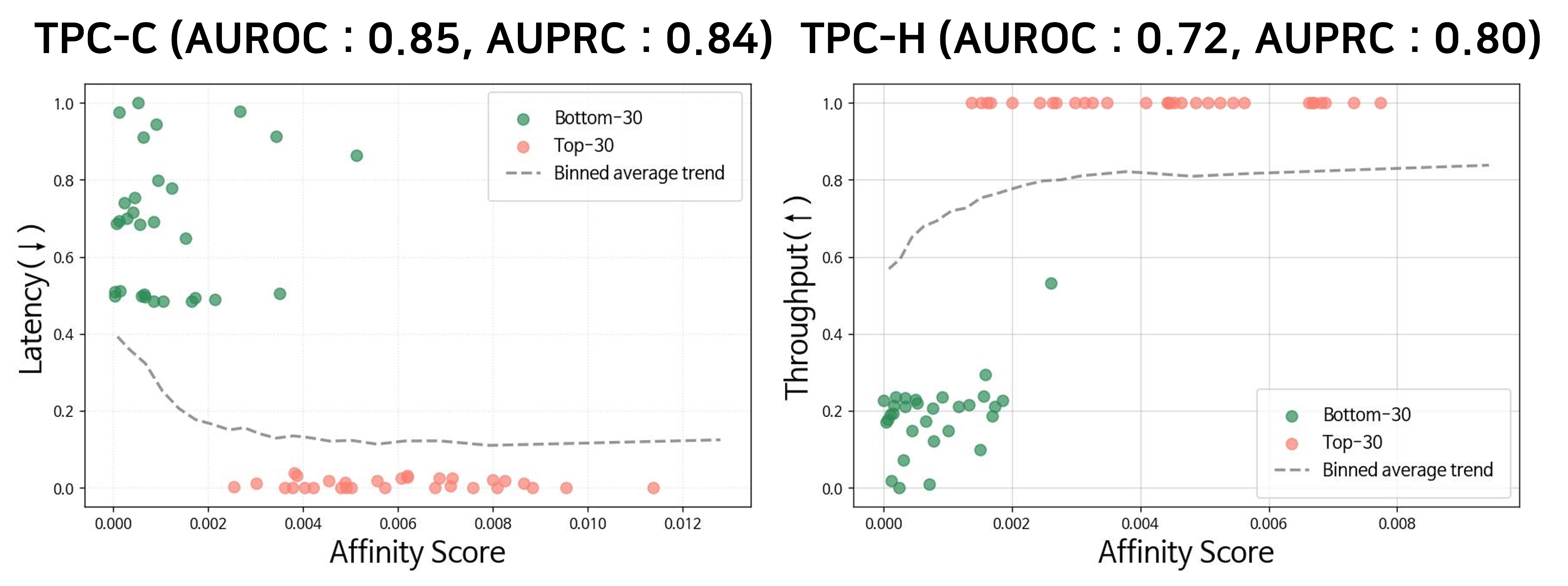}  
        \caption{Correlation between Affinity Score and Performance Metrics for PostgreSQL.}
        \label{fig:affinity_ex_post}
    \end{minipage}
\end{figure}

\noindent For the YCSB A workload, the binned trend clearly demonstrates that as the affinity score increases, throughput gradually rises while latency decreases, revealing a consistent overall trend. The affinity score exhibited a clear correlation with throughput and latency, showing AUROC = 0.82 and AUPRC = 0.88 in quantitative evaluation. Latent vectors with higher affinity values tended to achieve higher throughput and lower latency, indicating that the proposed affinity score effectively captures the performance-relevant structure within the latent space. Although a few latent vectors in the mid-range affinity intervals exhibit relatively lower performance, the overall results indicate that the Affinity Score reliably identifies high-performing regions within the latent space.

For the YCSB F workload, the correlation between the affinity score and performance was even more pronounced,
with AUROC = 0.98 and AUPRC = 0.99, indicating a very strong relationship. The relation between affinity and throughput exhibited an almost linear increasing pattern, where latent vectors with higher affinity values consistently achieved higher throughput and lower latency.

Figure \ref{fig:affinity_ex_post} shows the results for PostgreSQL on the TPC-C and TPC-H workloads. For the TPC-C workload, the relationship between latency and the affinity score exhibited a highly consistent monotonic decreasing pattern. Quantitatively, AUROC and AUPRC reached 0.85 and 0.84, respectively, demonstrating a strong negative correlation. As the affinity value increased, latency sharply decreased, forming a clear downward binned-average trend. This observation indicates that the affinity score effectively captures the sensitive variations in latency performance, serving as a reliable discriminative indicator within the latent space even for transaction-oriented workloads such as TPC-C.

For the TPC-H workload, despite its relatively low structural variability as an analytical query workload, the affinity score showed a meaningful positive correlation with throughput. AUROC and AUPRC were 0.72 and 0.80, respectively, revealing that latent vectors with higher affinity values tended to cluster in regions with higher throughput. The binned trend exhibited a gradually increasing shape, suggesting that the affinity score maintains a consistent level of predictive capability even under workloads with limited structural diversity.

Overall, the proposed affinity score exhibited a strong positive correlation with throughput and a strong negative correlation with latency, demonstrating consistent performance patterns across diverse workloads.
These results confirm that the affinity score serves as a reliable indicator of performance proximity within the latent space.

\begin{figure}[t]
  \centering

  \begin{subfigure}{\columnwidth}
    \centering
    \includegraphics[width=\linewidth]{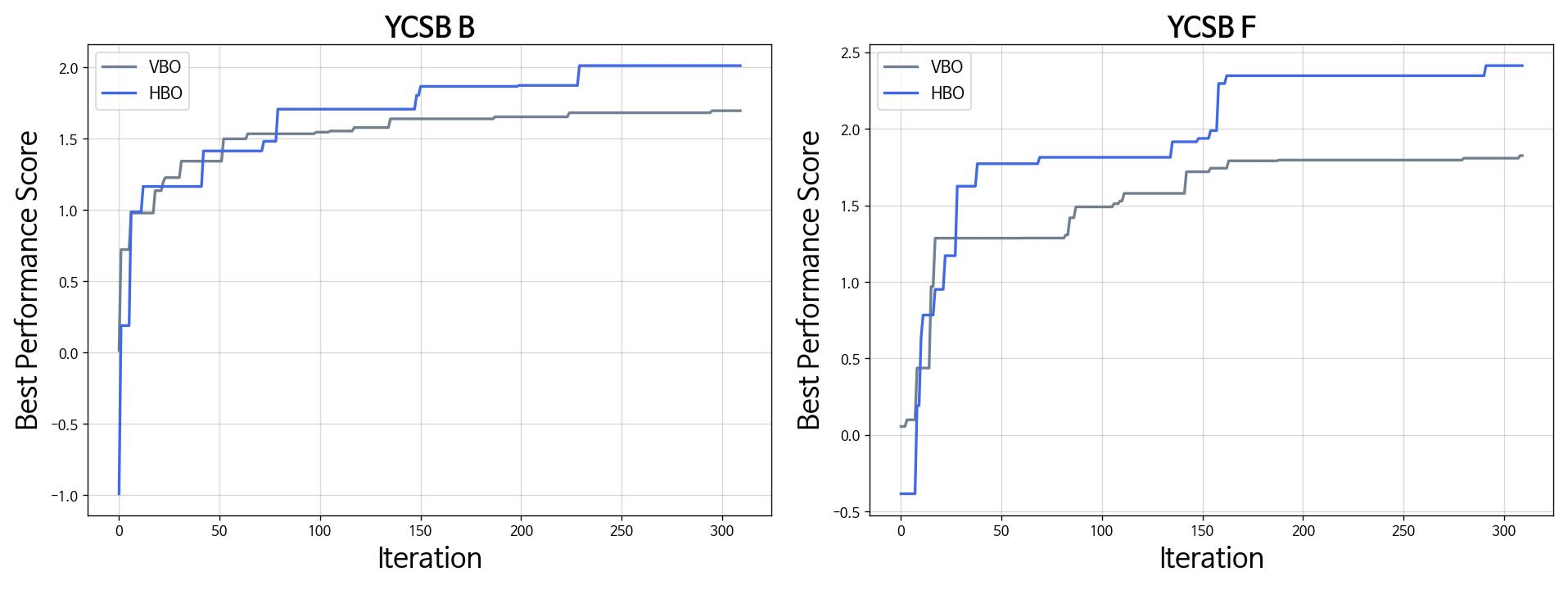}
    \subcaption{Convergence comparison between VBO and HBO for MySQL.}
    \label{fig:hbo_mysql}
  \end{subfigure}

  \medskip 

  \begin{subfigure}{\columnwidth}
    \centering
    \includegraphics[width=\linewidth]{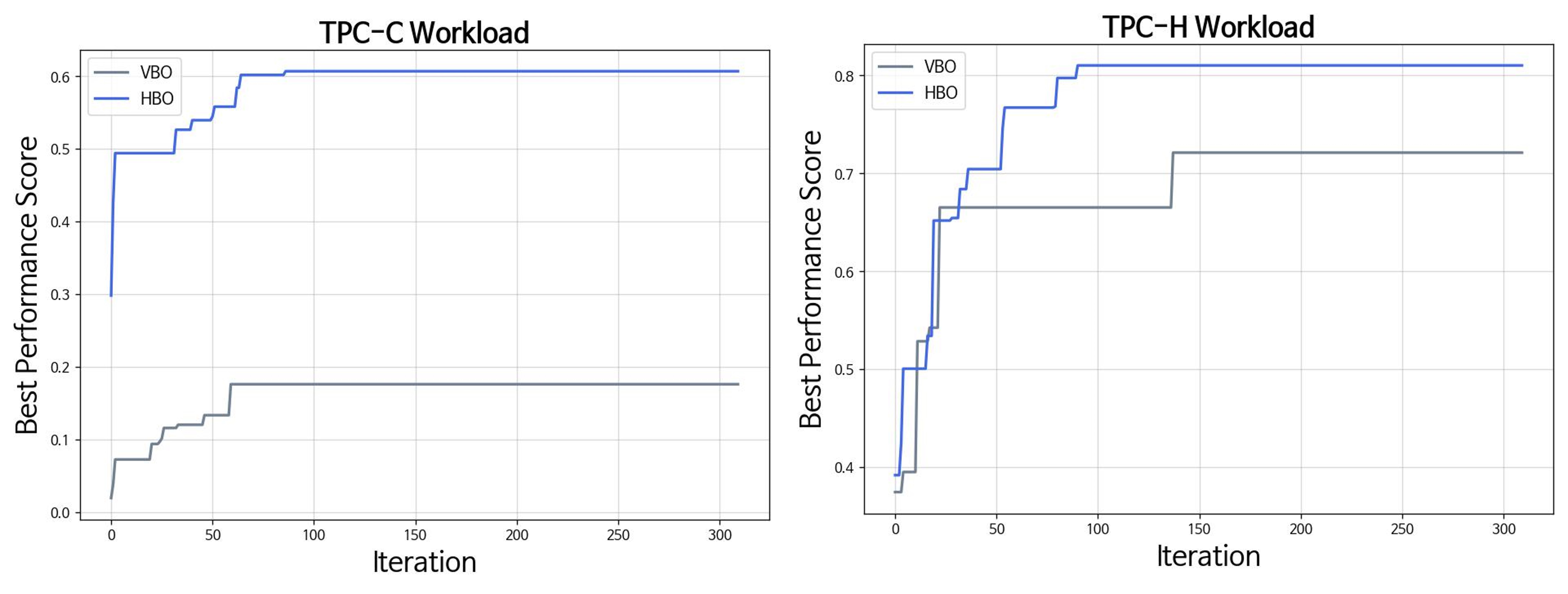}
    \subcaption{Convergence comparison between VBO and HBO for PostgreSQL.}
    \label{fig:hbo_post}
  \end{subfigure}

  \caption{Convergence comparisons on two DBMSs.}
  \label{fig:hbo_both}
\end{figure}

\subsection{Comparison between Vanilla BO and Hybrid-Score-Guided BO}
To evaluate the search efficiency of the proposed Hybrid-Score-Guided Bayesian Optimization (HBO), we compared its performance against the conventional Vanilla Bayesian Optimization (VBO). The comparison experiments were conducted on two MySQL workloads (YCSB B and YCSB F) and two PostgreSQL workloads (TPC-C and TPC-H). Each experiment was performed over 300 tuning iterations, and the progression of the best performance score was plotted to visualize the optimization trajectory.

As shown in Figure \ref{fig:hbo_mysql}, for the YCSB B and YCSB F workloads, HBO converges to high performance regions more rapidly than VBO within the first 100 iterations. In particular,for YCSB F, HBO achieves over a 30\% improvement in performance around 150 iterations and continues to maintain a consistently higher score subsequently. This demonstrates that by jointly considering both the surrogate model’s prediction and the Affinity Score, HBO more effectively guides the search toward already verified high performing regions within the latent space.

Figure \ref{fig:hbo_post} shows that TPC-C and TPC-H, HBO consistently outperforms VBO for PostgreSQL. In the case of TPC-C, HBO exhibited a sharp increase in score from the early iterations and converged near the optimum within 100 iterations, whereas VBO showed unstable fluctuations during the initial exploration phase and eventually stagnated at a lower performance level. This result suggests that in transaction oriented workloads, HBO effectively captures performance sensitive regions early and mitigates unnecessary local exploitation. For the TPC-H workload, although the overall score variation was limited due to the analytical nature of the workload, HBO reached high-performing regions faster than VBO and achieved a higher final performance level.

Overall, HBO consistently outperformed VBO in both search efficiency and convergence speed.
This improvement can be attributed to the Affinity-Score-guided search, which enables more effective identification of high-performing regions within the latent space.
These results demonstrate that HBO goes beyond conventional surrogate-model-based exploration by performing directional optimization that reflects the performance-relevant geometry of the latent space.


\begin{table}[t]
\caption{Time cost (minutes) to complete tuning across workloads. Lower is better.}
\label{tab:time_cost}
\centering
\small
\renewcommand{\arraystretch}{1.3}

\begin{tabular*}{\columnwidth}{@{\extracolsep{\fill}} l cccc @{}}
\toprule
\textbf{Method} & \textbf{YCSB B} & \textbf{YCSB F} & \textbf{TPC-C} & \textbf{TPC-H} \\
\midrule
OtterTune & \textbf{50.16} & 49.70 & 264.17 & 345.80 \\
CDBTune   & 370.15 & 421.85 & 462.25 & 424.61 \\
RGPE      & 68.46  & \textbf{34.24} & 197.57 & 224.70 \\
CSAT      & 242.45 & 267.53 & 428.35 & 467.20 \\
GPTuner   & 276.20 & 284.24 & 305.40 & 357.50 \\
RelTune   & 183.13 & 164.29 & \textbf{192.61} & \textbf{167.16} \\
\bottomrule
\end{tabular*}
\end{table}

\subsection{Analysis of Tuning Overhead and Efficiency}
Table \ref{tab:time_cost} compares the time cost required by each baseline to complete tuning across different workloads. For the MySQL YCSB workloads, RelTune exhibited a slightly longer tuning time due to the additional GAT training phase on the parameter relationship graph. This initial training step, which encodes parameter dependencies into a latent representation, introduces extra overhead compared to purely search-based approaches. Unlike traditional methods, RelTune requires an initial model training stage to encode these dependencies into a latent representation space. Nevertheless, RelTune achieved the highest performance on both YCSB B and YCSB F workloads, indicating that the initial learning cost is offset by improved exploration efficiency during the tuning process.

In contrast, for the PostgreSQL workloads (TPC-C and TPC-H), RelTune achieved the shortest tuning time among all baselines. This is because performance evaluation through actual benchmark execution is particularly time-consuming in PostgreSQL. Most baseline methods perform real DB restarts and benchmark executions for each BO iteration to measure TPS and latency, resulting in significant overhead throughout the optimization process. 
RelTune, on the other hand, as shown in line 10 of Algorithm \ref{alg:hbo}, does not execute the actual benchmark in every iteration. Instead, its surrogate model’s metric prediction head \text{$e_\psi$} evaluates performance based on the predicted latency in the latent space. By estimating performance through surrogate prediction rather than expensive real executions, RelTune substantially reduces the overall tuning time.

Furthermore, the benchmark execution in PostgreSQL is inherently slower than in MySQL, mainly due to its heavier transaction commit and Write-Ahead Logging (WAL) mechanisms, as well as the cluster re-initialization and data reloading steps that benchmark performs at each iteration. Even under such system-level constraints, RelTune effectively minimizes the number of actual executions through surrogate-based evaluation, achieving the highest time efficiency among all methods in the PostgreSQL environment.

\newcommand{\cmark}{\textcolor{green!60!black}{\ding{51}}}
\newcommand{\xmark}{\textcolor{red!70!black}{\ding{55}}}

\renewcommand{\arraystretch}{1.25}
\setlength{\tabcolsep}{6pt}

\begin{table}[t]
\centering
\caption{Ablation study results on MySQL workloads.}
\label{tab:ablation}

\begin{subtable}{0.9\columnwidth}  
\centering
\resizebox{\linewidth}{!}{%
\begin{tabular}{c|c|cc|cc}
\hline
\multicolumn{2}{c|}{\textbf{Module}} & \multicolumn{2}{c|}{\textbf{YCSB A}} & \multicolumn{2}{c}{\textbf{YCSB B}} \\ \hline
RGE & HBO & TPS & Latency & TPS & Latency \\ \hline
\xmark & \xmark & 3684.76 & 1426 & 9968.24 & 395 \\
\cmark & \xmark & 4071.56 & 1146 & 9967.39 & 353 \\
\xmark & \cmark & 3741.25 & 1169 & 9972.83 & 320 \\
\rowcolor{blue!10}
\cmark & \cmark & \textbf{4520.11} & \textbf{1028} & \textbf{9999.99} & \textbf{230} \\ \hline
\end{tabular}%
}
\end{subtable}

\vspace{0.8em} 

\begin{subtable}{0.9\columnwidth}
\centering
\resizebox{\linewidth}{!}{%
\begin{tabular}{c|c|cc|cc}
\hline
\multicolumn{2}{c|}{\textbf{Module}} & \multicolumn{2}{c|}{\textbf{YCSB E}} & \multicolumn{2}{c}{\textbf{YCSB F}} \\ \hline
RGE & HBO & TPS & Latency & TPS & Latency \\ \hline
\xmark & \xmark & 3862.43 & 1421 & 3357.27 & 1652 \\
\cmark & \xmark & 4774.22 & 1330 & 4071.56 & 1146 \\
\xmark & \cmark & 5319.82 & 1115 & 3400.37 & 1270 \\
\rowcolor{blue!10}
\cmark & \cmark & \textbf{5766.32} & \textbf{833} & \textbf{4520.11} & \textbf{1028} \\ \hline
\end{tabular}%
}
\end{subtable}
\end{table}

\subsection{Ablation Study}
Table \ref{tab:ablation} presents the ablation study results evaluating the contributions of the two core components of the proposed RelTune framework: the \textit{Relational Graph Encoder (RGE)} and the \textit{Hybrid-Score-Guided Bayesian Optimization (HBO)}. 

In the experiment without the RGE module, the relational graph in the GNN was removed, and a simple MLP encoder was used that takes only configuration–metric pairs as input. Under this setting, each parameter was treated as an independent feature when constructing the latent representation, thereby excluding dependency information among parameters. As a result, the expressiveness of the latent space was degraded, leading to a noticeable decline in both TPS and latency performance. These findings support that relational-aware encoding plays a crucial role in capturing inter-parameter dependencies within database configurations.

In the experiment without the HBO module, the affinity term in the hybrid score was removed, and a vanilla Bayesian Optimization (VBO) was used for optimization.
Although the search process remained functional, the absence of affinity-guided exploration resulted in lower performance compared to HBO under the same number of iterations.
This outcome supports that the hybrid score enables more efficient exploration by guiding the search toward latent vectors located near high-performing regions, allowing HBO to discover better-performing configurations faster than VBO.

As a result, when both modules were used together, RelTune achieved the highest TPS and the lowest latency across all workloads. These results demonstrate that generating latent vectors that incorporate inter-parameter relationships and optimizing those vectors through HBO work in a complementary manner to enhance overall performance.

\section{conclusion}
This study addresses three key limitations in existing database configuration tuning methods:
(1) the neglect of inter-parameter relationships,
(2) inefficient partial tuning to mitigate high-dimensional search space complexity, and
(3) the limited optimization efficiency of conventional search-based approaches. To overcome these issues, we propose RelTune, a framework that models parameter dependencies through a Relational Graph Representation and learns them via a Graph Neural Network (GNN) encoder–decoder architecture. 
This design produces latent representations that capture semantic and structural interactions among parameters, enabling more consistent and stable tuning behavior. 
We further introduce the Affinity Score, a quantitative measure of proximity to high-performing regions in the latent space, which shows strong correlation with throughput and latency metrics across workloads. 
By integrating this metric into Bayesian Optimization, our Hybrid-Score-Guided BO (HBO) achieves faster convergence and superior final performance compared to vanilla BO.

In summary, RelTune unifies relational structure modeling with efficient optimization, achieving high tuning efficiency across diverse DBMSs and workloads, and lays the foundation for future extensions toward real-time adaptive tuning.

\section*{Acknowledgment}
This research was supported by the National Research Foundation (NRF) funded by the Korean government (MSIT) (No. RS-2023-00229822).

\bibliographystyle{IEEEtran}
\bibliography{ref}

\end{document}